\documentclass{article} 
\usepackage{iclr2022_conference,times}

\usepackage{microtype}
\usepackage{graphicx}
\usepackage{subfigure}
\usepackage{booktabs} 
\usepackage{caption}
\usepackage{wrapfig}
\usepackage{algorithm}
\usepackage{algorithmic}
\usepackage{physics}
\usepackage{arydshln}

\newcommand\std[1]{\ensuremath{\color{gray} \scriptstyle \pm #1}}

\usepackage{amsmath,amsfonts,bm}

\def\eqref#1{equation~\ref{#1}}

\def\1{\bm{1}}

\DeclareMathAlphabet{\mathsfit}{\encodingdefault}{\sfdefault}{m}{sl}
\SetMathAlphabet{\mathsfit}{bold}{\encodingdefault}{\sfdefault}{bx}{n}

\usepackage{url}
\usepackage{multirow}
\usepackage{makecell}

\usepackage[colorlinks=true,linkcolor=blue,citecolor=blue]{hyperref}
\usepackage[skins,theorems]{tcolorbox}
\usepackage{colortbl}

\usepackage[disable,textsize=tiny]{todonotes}
\newcommand{\kensen}[1]{\todo[color=green!40]{kensen: #1}}

\newcommand{\pengcheng}[1]{\todo[color=blue!30]{PY: #1}}

\usepackage{appendix}
\usepackage{enumitem}

\def\programs{{\mathcal{P}}}
\def\specs{{\mathcal{X}}}
\newcommand{\logicalOR}{\; | \;}
\newcommand{\T}[1]{\texttt{#1}}
\newcommand{\redT}[1]{{\color{red!70!black}\texttt{#1}}}
\newcommand{\nl}[1]{{\color{green!60!black}``#1"}}
\newcommand{\actions}[1]{{\color{red!70!black}\texttt{[#1]}}}

\newcommand{\LengthGen}{{\color{blue!70!black}\emph{Length-Generalization}}}
\newcommand{\LengthGenHard}{{\color{blue!70!black}\emph{Length-Generalization-Hard}}}
\newcommand{\LengthGenHardest}{{\color{blue!70!black}\emph{Length-Generalization-Hardest}}}
\newcommand{\ComposeDiffConcepts}{{\color{blue!70!black}\emph{Compose-Different-Concepts}}}
\newcommand{\SwitchConceptOrder}{{\color{blue!70!black}\emph{Switch-Concept-Order}}}
\newcommand{\ComposeNewOp}{{\color{blue!70!black}\emph{Compose-New-Operation}}}
\newcommand{\AddOpFunctionality}{{\color{blue!70!black}\emph{Add-Operation-Functionality}}}

\title{Compositional Generalization and \\ Decomposition in Neural Program Synthesis}

\author{%
Kensen Shi \\
Google Research \\
\small\texttt{kshi@google.com}
\And
Joey Hong \\
UC Berkeley \\
\small\texttt{joey\_hong@berkeley.edu}
\And
Manzil Zaheer \\
Google Research \\
\small\texttt{manzilzaheer@google.com}
\And
Pengcheng Yin \\
Google Research \\
\small\texttt{pcyin@google.com}
\And
Charles Sutton \\
Google Research \\
\small\texttt{charlessutton@google.com}
}

\iclrfinalcopy 
\begin{document}

\maketitle


\begin{abstract}
When writing programs, people have the ability to tackle a new complex task by decomposing it into smaller and more familiar subtasks. While it is difficult to measure whether neural program synthesis methods have similar capabilities, what we can measure is whether they compositionally generalize, that is, whether a model that has been trained on the simpler subtasks is subsequently able to solve more complex tasks. In this paper, we focus on measuring the ability of learned program synthesizers to compositionally generalize. We first characterize several different axes along which program synthesis methods would be desired to generalize, e.g., length generalization, or the ability to combine known subroutines in new ways that do not occur in the training data. Based on this characterization, we introduce a benchmark suite of tasks to assess these abilities based on two popular existing datasets, SCAN and RobustFill. Finally, we make first attempts to improve the compositional generalization ability of Transformer models along these axes through novel attention mechanisms that draw inspiration from a human-like decomposition strategy. Empirically, we find our modified Transformer models generally perform better than natural baselines, but the tasks remain challenging.
\end{abstract}

\section{Introduction}
\label{sec:introduction}

\emph{Program synthesis} aims to assist programmers by automatically producing code according to a user's specification of what the code should do~\citep{gulwani2017program}.
Search-based program synthesis approaches, such as programming by example (PBE) systems, have been effective 
for small self-contained tasks such as short Java functions~\citep{shi2019frangel}, string manipulation \citep{FLASHFILL,Devlin2017Robustfill,CROSSBEAM}, 
and tensor manipulation~\citep{shi2020tf}.
However, synthesizing complex or long programs can be expensive because the search space grows exponentially with respect to the program length.
Furthermore, many search-based approaches require significant engineering effort to adapt to new programming libraries or languages.
Similarly, program synthesizers that use constraint solving \citep{SKETCH, ROSETTE} are successful in narrow domain-specific languages (DSLs), but they often become intractable for longer programs and can be difficult to extend beyond simple DSLs.
Neural program synthesizers, especially those based on large language models~\citep{chen2021evaluating,Austin2021-fu,ALPHACODE},
can produce longer or more complex code at a lower computation cost, but their successes are often limited to examples similar to those present in the training data~\citep{Furrer2020CompositionalGI}. That is, they do not generalize well to new APIs, novel concepts, or even novel \emph{combinations} of concepts.

It is desirable for program synthesizers to generalize in many ways. For example, an ideal synthesizer would produce longer code without a prohibitive increase in computational cost or a dramatic decrease in code quality. It would adapt its general programming skills to handle new APIs and concepts with minimal extra guidance or engineering. It would also be able to mix and match programming concepts, composing different code idioms in novel ways to solve novel problems. These are all \emph{compositional generalization} skills that human programmers naturally develop but are often difficult for program synthesizers. Compositional generality means the ability to generalize to test examples consisting of compositions of components seen during training, but where the distribution of compositional patterns is different~\citep{keysers20cfq}. While current program synthesizers are far from reaching these lofty goals, we can measure the compositional generalization abilities of different synthesis techniques to help push the state-of-the-art toward these desirable human-like abilities.

Prior work has evaluated whether natural language processing systems can compositionally generalize, proposing benchmark datasets 
to measure the ability of language understanding models in interpreting learned concepts, e.g., \textit{jump}, in compositionally novel contexts, e.g., \textit{jump twice} \citep{Marcus2001TheAM,Lake2018Scan}.
We adapt those ideas to focus on how problem-solving in the form of \emph{programming} is compositional. 
For instance, complex computer programs are typically built by composing individual functions and API calls, which can be composed in novel ways to solve novel problems.
In this paper, we identify seven different compositional generalization tasks applicable to program synthesis and propose a new method of creating benchmark datasets that measure these forms of compositional generalization, for both zero-shot and few-shot generalization.
We apply our benchmark-creation method to two popular domains: SCAN~\citep{Lake2018Scan}, which involves generating a sequence of actions specified by a natural language-like command, and RobustFill~\citep{Devlin2017Robustfill}, which targets string manipulation programs specified by input-output examples. Our benchmark-creation method is agnostic to the kind of program specification used, and our RobustFill-based benchmark is the first compositional generalization dataset using input-output examples to our knowledge, making it particularly applicable to program synthesis.

In addition to proposing benchmark datasets to measure the compositional generality of program synthesizers, we furthermore hypothesize that \emph{decomposition} is particularly useful for achieving this compositional generality. Decomposition is the problem-solving technique (broadly applicable even beyond programming) of breaking a complex task into multiple smaller parts, perhaps repeatedly, until each subtask is easy enough to handle. Decomposition is especially important within software engineering where implementations of subtasks can be combined and reused in modular ways. Applying decomposition is a skill so fundamental to software engineering that the first programming course at Stanford University begins teaching decomposition within the first week of class, immediately after introducing basic syntax, the concept of functions, and simple control flow~\citep{cs106a}. Because compositional generality revolves around combining ideas in new ways, and the decomposition approach solves problems by isolating subtasks and combining their solutions, we argue that a decompositional programming strategy is likely to have high compositional generality (although this is not necessarily the only viable strategy). Hence, we propose variations of the Transformer architecture motivated by the decomposition strategy, where the model is trained to recognize boundaries between subtasks and focus on solving one subtask at a time. As a bonus, well-decomposed code is a hallmark of good coding style, so it is additionally desirable to encourage synthesizers to produce such code.

In our experiments, we find that our decomposition-based Transformer variations outperform the vanilla Transformer architecture for some but not all of the compositional generalization tasks, with a greater improvement over the baseline for SCAN than for RobustFill. Even so, our compositional generalization benchmarks remain difficult overall in both the zero-shot and few-shot settings. We hope that the datasets inspire continued research using different kinds of techniques toward the goal of compositionally general program synthesis.

\section{Compositional Generalization in Programming}
\label{sec:generalization}

The goal in program synthesis is to find a program in a given language that is consistent with a specification.
Formally, we are given a domain specific language (DSL) which defines a space $\programs$ of programs. 
The task is described by a specification $X \in \specs$ and is solved by an unknown program $P^* \in \programs$.
For example, a specification can be a set of input/output (I/O) examples denoted $X = \{(I_1, O_1), \hdots (I_N, O_N)\}$. Then, solving specification $X$ means finding a program $P$ (not necessarily $P^*$) that correctly solves all of the examples: $P(I_i) = O_i, \, \forall i$. A specification can also be a natural language description of a task, and the corresponding program implements said task.

Program synthesizers are more robust and more broadly applicable if they generalize well. In this section we discuss several distinct forms of generalization that are desirable in program synthesis. Current program synthesizers do not achieve high generalization in all of these ways.
At a high level, we identify three broad categories of generalization that an ``ideal program synthesizer'' should have:
\begin{itemize}[leftmargin=1em]
  \item \emph{Length generalization}: Produce longer code than appeared in the training set, but without a prohibitive increase in computational cost, and without a substantial decrease in quality.
  \todo{CS: P1, we could rearrange this section to integrate this list with the list of seven. This might break up the seven in a way that makes it easier to read.}
  
    
    \item \emph{Mix and match concepts}: Compose code idioms in novel ways to solve novel problems, but without combinatorially many training examples covering all combinations.
    
    
    \item \emph{Apply general principles}: Adapt to new, updated, or custom APIs by drawing on knowledge of other similar APIs, without excessive guidance or engineering.
    \todo{This one feels too  vague to me, would we be happy with: Few-shot concept learning: From a small number of examples of a new API method, correctly combine it with already-known programming constructs to solve new problems?}
    \pengcheng{I feel what I wrote is just one instnace of ``apply general principles'' (update functions). Shall we make it more abstract here?}
    \kensen{I made the original text more specific, but still less specific than Charles and Pengcheng's suggestions. I think it's ok for this ``broad category'' of generalization to be broader than our specific tasks. E.g., it makes sense for this category to include ``understanding TensorFlow helps with understanding PyTorch'' which is too complex for us to actually turn into a concrete compositional generalization task.}
\end{itemize}

These kinds of generalization can be described as \emph{compositional generalization}, which revolves around understanding how basic building blocks can be composed in different ways to create larger structures. Prior work in natural language processing has studied compositionality in natural language~\citep{chomsky2002syntactic,talmor2018web,Lake2018Scan,keysers20cfq,gu2021beyond}.
For example, the SCAN dataset~\citep{Lake2018Scan} tests compositional generalization for translation models. The SCAN task is to translate from a language-like command such as \nl{jump left twice and walk} to a sequence of actions, in this case \actions{LTURN, JUMP, LTURN, JUMP, WALK}. One compositional generalization task would be to train a model on commands except for those including \nl{jump right}, and then test on commands containing \nl{jump right}. This zero-shot generalization task requires the model to understand \nl{jump} and \nl{right} individually, as well as the compositional pattern of an action verb followed by a direction. Such understanding can be drawn from other training examples including \nl{jump left}, \nl{walk left}, and \nl{walk right}.\todo{CS P1: This discussion of SCAN can be compressed to 1 sentence, because we will discuss SCAN in detail later.}

We adapt compositional generalization to the context of program synthesis, focusing on how problem-solving (in the form of programs) is compositional. Regardless of the programming language or DSL, programs nearly always consist of compositions of smaller parts. In general, a ``program part'' could mean a line of code, a block of statements, a function, or some other natural DSL-specific portion of code, where multiple such portions can be combined into a full program. We can even view SCAN action sequences as programs (with SCAN commands being the specification). For SCAN, we choose to define ``parts'' to be the portions of the action sequence that were separated by conjunctions (\nl{and} and \nl{after}) in the command. In the example above, the program parts would be \actions{LTURN, JUMP, LTURN, JUMP} and \actions{WALK}, corresponding to \nl{jump left twice} and \nl{walk} from the command. We use the notion of program parts to study how they are combined with different compositional patterns. Thus, the conventions for partitioning a program into its composed parts may vary depending on the DSL and the compositional patterns being tested.

We expand the three broad categories of generalization above into 7 concrete compositional generalization tasks applicable to program synthesis. Each generalization task describes a method of creating training and test sets with disjoint distributions, so that generalization may be tested in a zero-shot or few-shot setting. The generalization tasks are as follows:
\begin{enumerate}[leftmargin=1.2em]

\item \textbf{\LengthGen}: 
Can a model \emph{produce longer code} than seen in training, when the problem requires it?
Here, ``length'' is measured by the number of composed parts in the program, not simply the number of tokens, so there is more emphasis on generalizing to more complex compositional patterns. For this generalization task, we train on problems of lengths 1 to $n$ and test on lengths $n+1$ to $m$ (where $m > n$). In our experiments we choose $n=6$ and $m=10$.

\item \textbf{\LengthGenHard}:
Similar to above, but train on problems of length exactly $n$ and test on lengths $1$ to $m$ except $n$. To succeed on this generalization task, the synthesizer must recognize that problems may have varying difficulty with corresponding solutions of varying lengths, without seeing this fact demonstrated during training.

\item \textbf{\LengthGenHardest}:
Similar to above, but train on tasks of length exactly $1$ and test on lengths $2$ to $n$. Because the training data has no examples of how program parts may be composed, we do not expect neural models to achieve high zero-shot generalization of this kind.
Few-shot generalization is more interesting for this task---after mastering how to solve individual parts of a program, the synthesizer must quickly learn the compositional patterns for composing those parts into larger programs.

\item \textbf{\ComposeDiffConcepts} (a form of ``mix and match concepts''):
Can a model \emph{use concepts in different combinations} than seen in training?
We partition the DSL operations into multiple groups or \emph{concepts},\footnote{Ideally, operations within a group should have meaningful commonalities that form one concept, and each concept should have roughly equal semantic complexity, but these are not strictly required.} train on compositions of operations from the same concept, and test on compositions from different concepts. For example, if two concepts consist of operations $\{A_1, A_2, \ldots\}$ and $\{B_1, B_2, \dots\}$ respectively, then this generalization task involves training on programs of the forms $A_i \,\circ\, A_j$ and $B_i \,\circ\, B_j$, and testing on the forms $A_i \,\circ\, B_j$ and $B_i \,\circ\, A_j$. As a real-world example, this generalization task is similar to training on scripts containing only TensorFlow or only NumPy, but synthesizing code using both libraries.
\pengcheng{One potential (open) question a reviewer might ask is whether it is always safe to apply such generalization, for example, what if $B_j$ cannot be used together with $A_i$?}
\kensen{That question is the reason I (initially) wrote down the required DSL properties for these tasks, which we don't have space to include}

\item \textbf{\SwitchConceptOrder} (a form of ``mix and match concepts''):
Can a model \emph{compose concepts in different orders} than seen in training?
We again partition the DSL operations into multiple concepts (groups). We train on compositions of operations drawn from one sequence of concepts and test on a different sequence of concepts, e.g., train on $A_i \,\circ\, B_j$ and test on $B_i \,\circ\, A_j$. As a real-world example, in the training data a function might be primarily used to validate inputs at the beginning of the code, but we want to use the function in a different context, e.g., to validate results at the end of the code.

\item \textbf{\ComposeNewOp} (a form of ``apply general principles''):
Can a model learn to \emph{use a new isolated operation within a larger composition}?
In this task, we train on the isolated operation and compositions without the operation, and test on compositions using the operation.
For instance, in the SCAN domain, we could train on \nl{walk left after run twice} and \nl{jump}, and test on \nl{jump left after jump twice}. A real-world example of this kind of generalization would be composing a new function with others in a larger solution, after seeing examples of the function used in isolation.

\item \textbf{\AddOpFunctionality} (a form of ``apply general principles''):
Can a model \emph{extend its understanding of an operation by drawing on parallels} to other operations?
We omit from the training data some functionality of an operation that could be inferred from other context, and test on programs using that functionality.
For instance, in the SCAN domain, we could train on commands that do not contain \nl{around right} (but contain other similar constructions like \nl{opposite left} and \nl{opposite right}), and test on commands containing \nl{around right}. This task can occur in the real world when a library function is upgraded with a new parameter whose behavior can be inferred from other functions.

\end{enumerate}


\section{Benchmark Creation}
\label{sec:benchmarks}

\pengcheng{This does not need to be resolved soon, but for future versions it might be a good idea to combine Sections 2 and 3. We can briefly introduce the two domains in Section 2 (after introducing some representative generalization patterns in programming using SCAN), and have a big table listing examples for each of the 7 generalization tasks on the two doamins, similiar to Table 1 in https://arxiv.org/pdf/2010.05465.pdf.}
\kensen{if we do that, i'd still like to have a separation between general ideas that can be applied to lots of domains, and our specific choices for our specific domains}
\pengcheng{Yes, definitely! I agree it's important to introduce the general compositional generalization tasks in problem solving before diving into those concrete instantiations.}

We create benchmark datasets for the 7 kinds of compositional generalization tasks described in \autoref{sec:generalization} for two popular domains, SCAN~\citep{Lake2018Scan} and RobustFill~\citep{Devlin2017Robustfill}. While \autoref{sec:generalization} focused on general descriptions of the tasks, this section instantiates these tasks for our specific domains.

Our benchmark creation process works for both natural-language specifications (as in SCAN) and input-output (I/O) specifications (as in RobustFill). While the SCAN domain was used in prior work in natural language processing~\citep{Lake2018Scan}, we have expanded the domain to make our benchmark more applicable to program synthesis. Furthermore, our RobustFill benchmark is the first dataset for measuring compositional generalization for program synthesis from I/O examples.

In the SCAN domain, the objective is to translate from a natural-language command to a program that is a sequence of actions. \citet{Lake2018Scan} originally describes SCAN commands as having at most one \nl{and} or \nl{after} conjunction, but we generalize the domain so that commands can contain an arbitrary number of \nl{and} and \nl{after} conjunctions between parts of the command.\footnote{To eliminate ambiguity in the correct ordering of parts, we say that \nl{and} has higher precedence than \nl{after}. For example, \nl{jump and run after walk} should be translated to \actions{WALK, JUMP, RUN}, and \emph{not} \actions{JUMP, WALK, RUN}.}
We treat these conjunctions as the boundaries between program parts, so a command with $n$ parts will have $n - 1$ conjunctions and should be translated to an action sequence containing $n$ corresponding parts, although corresponding command and action sequence parts will appear in different orders whenever there is an \nl{after} conjunction.
We show the DSL for commands, as well as how commands are translated to programs in \autoref{app:dsls}.

In the RobustFill domain, the objective is to synthesize a string manipulation program from I/O examples. A RobustFill program is a concatenation of expressions, where an expression may be an operation that extracts a substring from the input, an operation that returns a modified version of the input, a special Compose operation (applying a modification operation to the result of another operation),\todo{Is compose just function composition? If so, then the previous descriptions seems confusing to me. Also, I do not see compose in the appendix.}
\kensen{compose in the DSL is just $m_1(m_2)$ and $m(s)$, which is actually composing functions. When we say \emph{program parts} are "composed" for RobustFill, that just means the top-level concatenation. It is indeed confusing}
or a constant string. See \autoref{app:dsls} for the full DSL of how programs are generated. Due to this program structure, we treat each expression as a \emph{program part}.

\autoref{app:benchmark_details} provides more details for the setup of specific generalization tasks for both domains.

\section{Models}
\label{sec:models}

We approach our compositional synthesis benchmarks using a sequence-to-sequence (seq2seq) model, which has been shown to be successful on various natural language \citep{Attention,Transformer} and program synthesis tasks \citep{parisotto2017neuro,Devlin2017Robustfill}.
In this paper, we choose our seq2seq model to be a Transformer due to its impressive performance on natural language tasks over traditional RNNs \citep{Transformer}. 
\autoref{sec:baseline_transformer} describes a baseline Transformer adapted to program synthesis. In \autoref{sec:decompositional_transformer}, we present modifications to the baseline model to encourage decomposition. We call the modified architecture the \emph{Decompositional Transformer}.

\subsection{Baseline Transformer}
\label{sec:baseline_transformer}
Our baseline Transformer consists of two modules. First, a Transformer encoder receives the specification $X$ word-by-word and produces an encoding, $E \leftarrow \mathrm{TransformerEncoder}(X)$.
Then, a Transformer decoder takes the encoding and autoregressively generates a program token-by-token.
Formally, let $P_{t-1} = [p_1, p_2, \hdots, p_{t-1}]$ be the program generated so far. The decoder predicts the next program token as
$p_t \leftarrow \mathrm{TransformerDecoder}(P_{t-1}, E)$.
As described by~\citet{Transformer}, the Transformer encoder and decoder both apply a stack of self-attention and feed-forward units.

In the case of specification $X$ being multiple I/O examples, our Transformer architecture performs \emph{double attention} analogous to \citet{Devlin2017Robustfill}.
That is, for each example $(I_i, O_i)$, the encoder behaves as $E_i \leftarrow \mathrm{TransformerEncoder}(I_i, O_i)$, where the encoder now performs self-attention on input $I_i$ followed by cross-attention on output $O_i$ to $I_i$. Finally, the encoding $E$ is simply the concatenation across examples $E \leftarrow \mathrm{Concat}((E_i)_{i=1}^N)$ where $N$ is the number of examples.

\paragraph{Relative attention.} Early self-attention mechanisms have added representations of the absolute positions of tokens to its inputs \citep{Transformer}. However, we use representations of relative positions, or distances between tokens, in line with recent work showing that relative attention is advantageous, particularly on length-generalization tasks \citep{shaw2018relative,Csordas2021-ey}.
By considering logarithmic distances, our model is also encouraged to attend to more recent tokens during decoding, which can be desirable when programs consist of multiple smaller parts. 

\subsection{Decompositional Transformer}
\label{sec:decompositional_transformer}

Inspired by the human problem-solving and programming strategy of decomposition, our Decompositional Transformer architecture leverages the baseline architecture but ensures that program parts are decoded independently of one another, using novel attention mechanisms.
We accomplish this decomposition using two notable modifications to the baseline Transformer architecture:

\paragraph{Subprogram separators.}
We introduce a new separator token \T{SEP} to the program vocabulary.\footnote{In practice, we found it sufficient to have the \T{SEP} token be the same \T{BOS} token that marks the beginning of programs. In this manner, we avoid introducing a new token to the vocabulary.}
This token will partition programs into sequences of \emph{program parts}. The program parts can be composed, for instance via concatenation, to form the full program. We can straightforwardly add such tokens into the training data generated as described in \autoref{sec:benchmarks}. In this manner, we provide explicit training supervision for the compositional nature of the ground-truth programs.
Thus, our models are trained to predict a \T{SEP} token after completing each program part. To compensate during evaluation, we remove all \T{SEP} tokens from the generated program before evaluating its correctness.

\begin{figure*}[t]
\centering
\begin{minipage}{\textwidth}
    \centering
    \begin{tabular}{l l}
    Specification: & \nl{jump left twice and run right after walk thrice} \\[0.2em]
    Program: & \redT{WALK WALK WALK LTURN JUMP LTURN JUMP RTURN RUN} \\
    \end{tabular}
\end{minipage}
\begin{minipage}{\textwidth}
\vspace{0.1in}
\begin{minipage}{0.33\textwidth}
\includegraphics[width=\linewidth]{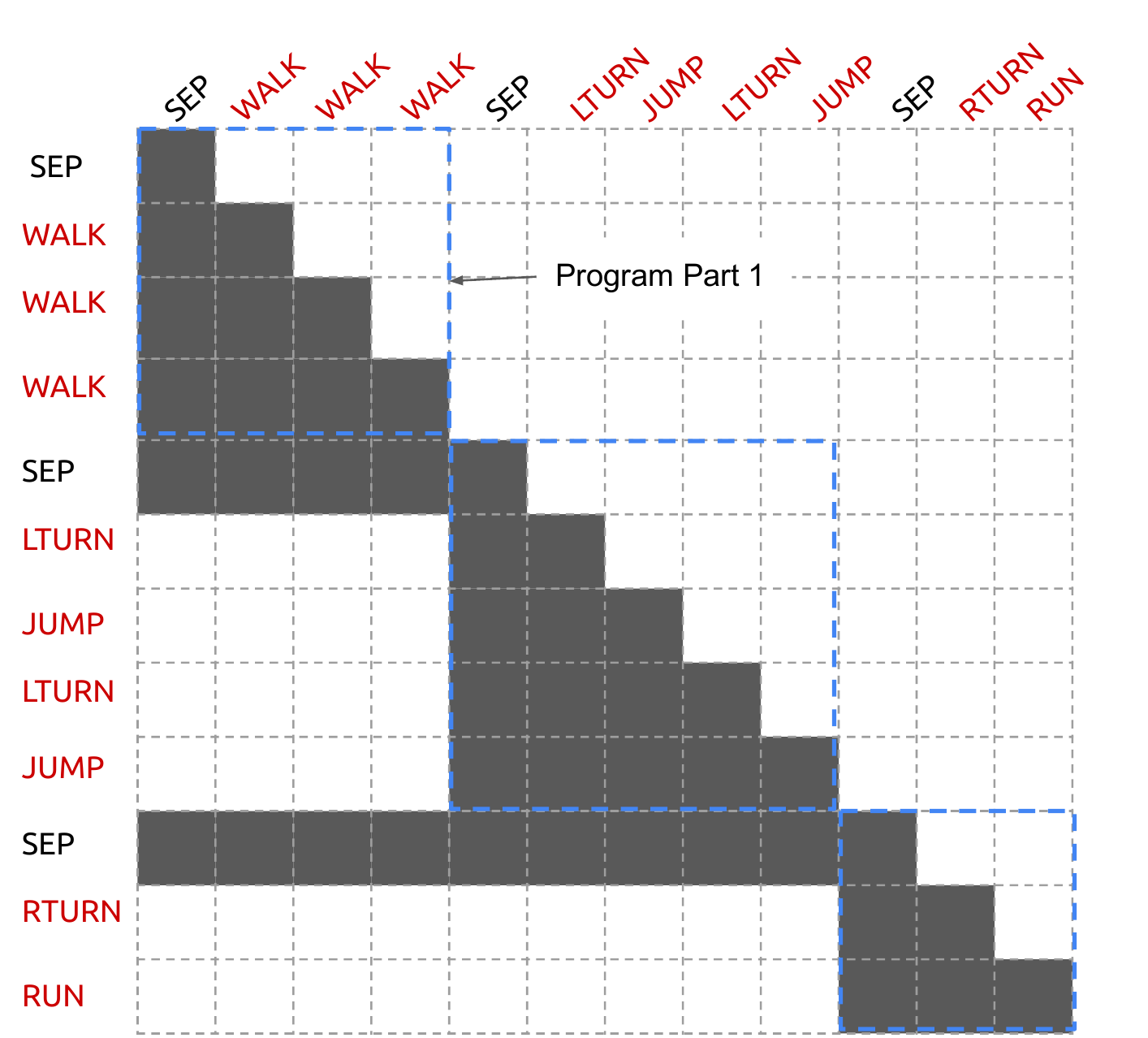}
\caption*{(a) Sep-Full-Attention}
\end{minipage}
\begin{minipage}{0.33\textwidth}
\includegraphics[width=\linewidth]{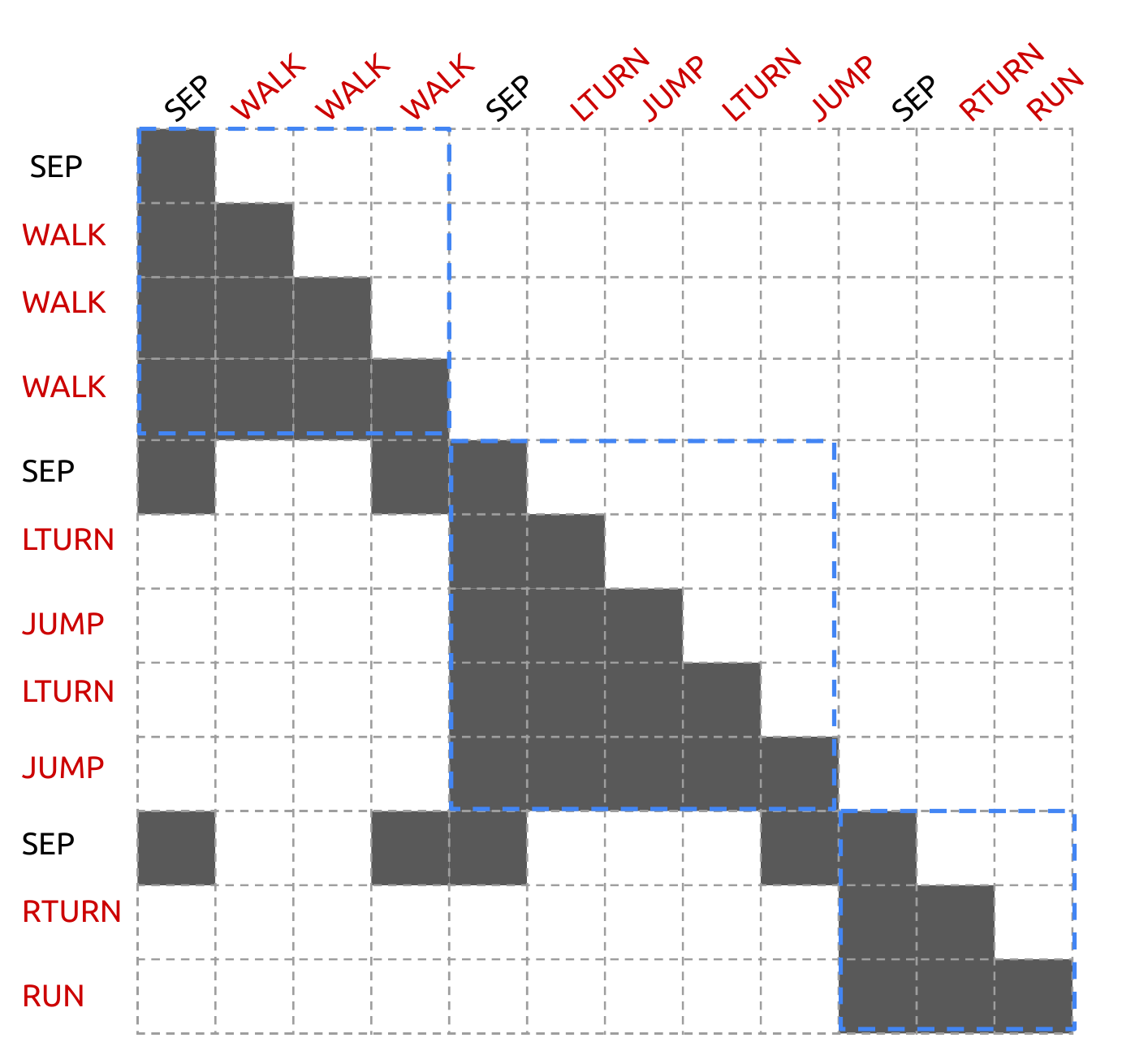}
\caption*{(b) Sep-to-Sep-and-Last}
\end{minipage}
\begin{minipage}{0.33\textwidth}
\includegraphics[width=\linewidth]{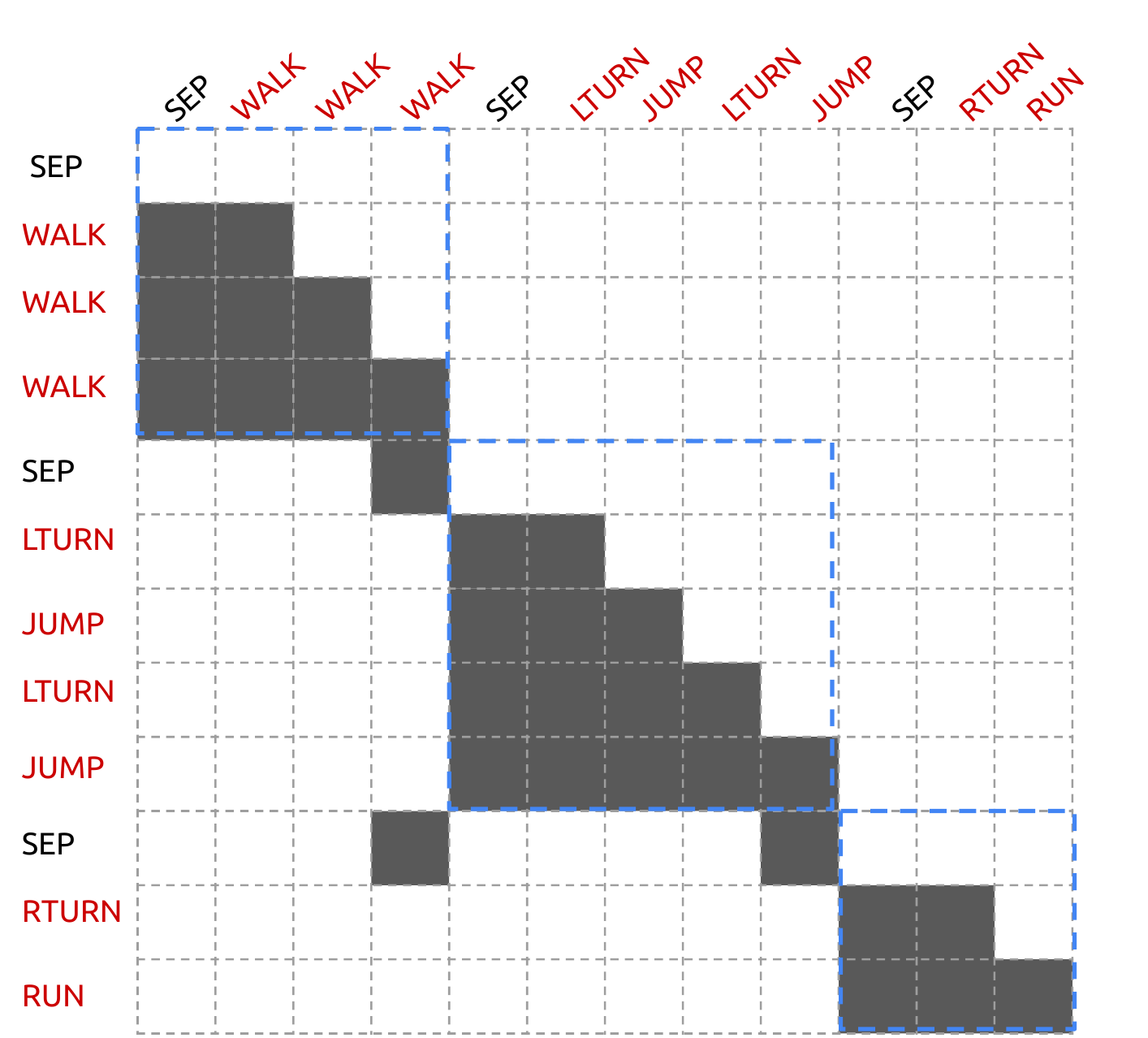}
\caption*{(c) Sep-to-Last}
\end{minipage}
\end{minipage}
\caption{Illustrations of our proposed attention mechanisms on an example SCAN program.}
\label{fig:attention_mask}
\end{figure*}

\paragraph{Decompositional attention masks.}
We incorporate novel attention mechanisms to ensure that program parts are decoded separately. As in the baseline architecture, we first use a Transformer encoder to produce an encoding $E \leftarrow \mathrm{TransformerEncoder}(X)$.
Let $P_{t-1}$ again be the program generated so far. 
In contrast to the baseline Transformer, during decoding, the next token in the program is instead generated as
$p_t \leftarrow \mathrm{TransformerDecoder}(Q_{t-1}, E)$,
where we apply an attention mask to the self-attention layers so that $Q_{t-1} \subseteq P_{t-1}$ consists of tokens relevant to solving the current subtask. More specifically, if $p_{t-1}$ is not \T{SEP}, then $Q_{t-1}$ consists of the program part generated so far, namely all program tokens starting from the most recent \T{SEP} (inclusive). Now, if $p_{t-1}$ is \T{SEP}, then our model must identify the next subtask to solve, which naturally should depend on the previously-solved subtasks.
We propose three different choices of $Q_{t-1}$ when $p_{t-1}$ is \T{SEP}, where the more tokens $Q_{t-1}$ contains, the more information we give to our model to identify the next subtask:
\begin{enumerate}[leftmargin=1.2em]
    \item \emph{Sep-Full-Attention}: In the most general case, we provide the Transformer decoder with the entire program generated so far, i.e., $Q_{t-1} = P_{t-1}$ when $p_{t-1}$ is \T{SEP}.
    \item \emph{Sep-to-Sep-and-Last}: In this case, $Q_{t-1}$ contains all previous \T{SEP} tokens, as well as the last program token in each previous program part. We use the last tokens because they attend to their entire respective program part during decoding, thus providing a summary of what that part does.
    \item \emph{Sep-to-Last}: In the most restrictive case, $Q_{t-1}$ only contains the last program token in each previously generated program part.
\end{enumerate}
We provide an illustration of each attention mask on an example program in \autoref{fig:attention_mask}. Note that relative attention encourages a similar behavior in attending more strongly to recent tokens; however, our attention masks are stricter and explicitly constrain attention to only include the necessary tokens.

\section{Experiments and Discussion}
\label{sec:experiments}

We trained the various methods in \autoref{sec:models} on each compositional generalization task in the SCAN and RobustFill datasets described in \autoref{sec:benchmarks}. We trained with batch size 128 and default settings for hyperparameters (details in \autoref{app:hyperparams}), training each method 3 times with different random initializations. We used 10K training steps for SCAN and 1M training steps for RobustFill, and we generated enough synthetic training data to avoid repeating examples. After training for each generalization task, we evaluate on 10K test examples to measure zero-shot generalization. Then, we fine-tune the trained models on a single batch containing 20 examples from the training distribution and 20 from the test distribution, repeated for 30 steps. Finally, we evaluate again on the same 10K test examples to measure few-shot generalization. SCAN results are in \autoref{tab:scan} and RobustFill results are in \autoref{tab:robustfill}. For a more visual representation, \autoref{app:plots} contains bar graphs for the same data.

All three improvements to vanilla Transformer (relative attention, separator tokens, and attention masks) generally help overall. In the most extreme case, our best model using the Sep-to-Last attention mask has 74.4\% zero-shot generalization for SCAN \ComposeDiffConcepts, compared to 29.2\% for a baseline Transformer with relative attention, or 10.5\% without any of the improvements. However, for some generalization tasks, the performance difference between models is much smaller.

Interestingly, adding separator tokens to the baseline Transformer (with and without relative attention) slightly decreases performance on zero-shot length generalization for both datasets.
This may be because the number of program parts is more obvious to the model, so the model is less likely to predict more \T{SEP} tokens than it has seen during training. Thus, the model may have difficulty generalizing to out-of-distribution lengths.
Despite this drawback, applying our attention masks (enabled by the separator tokens) leads to the best length generalization in most cases.
\begin{table}[t!]
\small
\centering
\setlength{\tabcolsep}{4.5pt}
\begin{tabular}{@{}ll@{}ccccccc@{}}
\toprule
    & & \multicolumn{3}{c}{Length Generalization}
    & \multirow{3}{1cm}[.3em]{\centering \footnotesize Compose Diff. Concepts}
    & \multirow{3}{1.1cm}[.3em]{\centering \footnotesize Switch Concept Order} 
    & \multirow{3}{1.1cm}[.3em]{\centering \footnotesize Compose New Operation} 
    & \multirow{3}{1.3cm}[.3em]{\centering \footnotesize Add Operation Func.} \\
    \cmidrule{3-5}
    & & 1-6 to 7-10    & 6 to 1-10   & 1 to 2-6        \\ 
\midrule
\multirow{7}{*}{\rotatebox{90}{Zero-shot}}
& \color{gray} Transformer & 26.9 \std{0.7} & 36.3 \std{0.6} & 1.1 \std{0.2} & 10.5 \std{2.4} & 4.6 \std{2.5} & 7.9 \std{2.2} & 41.7 \std{1.6} \\
& ~~$+$ Separators $\star$ & 26.7 \std{0.6} & 23.6 \std{0.6} & 0.9 \std{0.2} & 37.3 \std{2.8} & 10.0 \std{3.8} & 11.4 \std{1.1} & 43.7 \std{1.7} \\
& \color{gray} Relative Attention & 46.1 \std{3.0} & 39.8 \std{1.8} & 1.1 \std{0.1} & 29.2 \std{4.6} & 9.4 \std{4.0} & 9.3 \std{1.9} & 15.6 \std{6.8} \\
& ~~$+$ Separators $\star$ & 42.1 \std{3.5} & 33.8 \std{1.9} & 1.1 \std{0.1} & 46.7 \std{4.3} & 9.8 \std{4.8} & \textbf{12.9} \std{2.6} & 46.6 \std{7.0} \\
& Sep-Full-Attention $\star$ & \textbf{50.2} \std{2.0} & \textbf{41.2} \std{0.6} & 1.1 \std{0.1} & 36.7 \std{2.4} & 7.6 \std{5.5} & 11.6 \std{2.5} & 59.8 \std{4.4} \\
& Sep-to-Sep-and-Last $\star$ & 47.2 \std{2.5} & 34.4 \std{1.2} & \textbf{1.2} \std{0.1} & 49.0 \std{6.6} & 15.4 \std{4.0} & 10.3 \std{1.2} & \textbf{65.5} \std{15.1} \\
& Sep-to-Last $\star$ & 41.9 \std{3.3} & 24.1 \std{1.9} & 1.0 \std{0.2} & \textbf{74.4} \std{9.3} & \textbf{17.5} \std{8.8} & 12.5 \std{2.5} & 39.8 \std{6.8} \\
\midrule

\multirow{7}{*}{\rotatebox{90}{Fine-tuning}}
& \color{gray} Transformer & 28.7 \std{1.7} & 61.0 \std{1.6} & 1.1 \std{0.2} & 56.5 \std{6.4} & 48.3 \std{2.4} & 74.8 \std{0.6} & 89.6 \std{1.6} \\
& ~~$+$ Separators $\star$ & 32.6 \std{2.2} & 59.5 \std{3.6} & 1.0 \std{0.2} & \textbf{89.3} \std{1.7} & 58.4 \std{4.8} & 88.0 \std{1.0} & 97.6 \std{0.1} \\
& \color{gray} Relative Attention & 65.3 \std{1.1} & 67.5 \std{6.4} & 0.9 \std{0.3} & 53.0 \std{3.8} & 54.2 \std{4.5} & 79.4 \std{2.8} & 85.6 \std{3.6} \\
& ~~$+$ Separators $\star$ & 74.5 \std{6.9} & 71.2 \std{3.5} & 1.2 \std{0.2} & 76.4 \std{6.3} & 65.7 \std{1.0} & 89.0 \std{2.4} & 98.9 \std{0.6} \\
& Sep-Full-Attention $\star$ & 72.3 \std{2.8} & \textbf{76.7} \std{1.4} & \textbf{1.6} \std{0.1} & 71.2 \std{9.3} & 64.4 \std{3.1} & 88.9 \std{3.1} & 99.8 \std{0.1} \\
& Sep-to-Sep-and-Last $\star$ & \textbf{75.0} \std{2.3} & 71.5 \std{2.8} & 1.2 \std{0.1} & 72.3 \std{8.8} & 66.7 \std{3.6} & 93.2 \std{1.0} & 99.8 \std{0.0} \\
& Sep-to-Last $\star$ & 70.8 \std{4.5} & 69.5 \std{2.0} & 1.5 \std{0.2} & 84.9 \std{5.2} & \textbf{66.8} \std{1.5} & \textbf{95.7} \std{0.6} & \textbf{100.0} \std{0.0} \\
\bottomrule
\end{tabular}
\vspace{-2mm}
\caption{\textbf{SCAN} zero-shot generalization and few-shot fine-tuning results, with {\color{gray}$\pm\sigma$} denoting a standard deviation of $\sigma$ over 3 trials. Methods marked with $\star$ are newly proposed in this work.}
\label{tab:scan}
\end{table}

\begin{table}[t!]
\small
\centering
\setlength{\tabcolsep}{4.5pt}
\begin{tabular}{@{}ll@{}ccccccc@{}}
\toprule
    & & \multicolumn{3}{c}{Length Generalization}
    & \multirow{3}{1cm}[.3em]{\centering \footnotesize Compose Diff. Concepts}
    & \multirow{3}{1.1cm}[.3em]{\centering \footnotesize Switch Concept Order} 
    & \multirow{3}{1.1cm}[.3em]{\centering \footnotesize Compose New Operation} 
    & \multirow{3}{1.3cm}[.3em]{\centering \footnotesize Add Operation Func.} \\
    \cmidrule{3-5}
    & & 1-6 to 7-10    & 6 to 1-10   & 1 to 2-6        \\ 
\midrule
\multirow{7}{*}{\rotatebox{90}{Zero-shot}}
& \color{gray} Transformer & 39.1 \std{1.2} & 28.7 \std{1.0} & 1.8 \std{0.0} & 51.3 \std{0.5} & 4.7 \std{0.3} & 45.9 \std{0.1} & 55.3 \std{0.2} \\
& ~~$+$ Separators $\star$ & 28.5 \std{2.4} & 24.7 \std{0.6} & 1.8 \std{0.0} & 55.7 \std{1.7} & \textbf{6.6} \std{0.5} & 46.2 \std{0.1} & 55.6 \std{0.2} \\
& \color{gray} Relative Attention & 43.6 \std{3.0} & \textbf{30.8} \std{0.8} & 1.8 \std{0.0} & 56.1 \std{2.6} & 6.5 \std{0.5} & 46.2 \std{0.1} & 55.9 \std{0.2} \\
& ~~$+$ Separators $\star$ & 43.5 \std{3.0} & 29.5 \std{2.3} & \textbf{1.9} \std{0.0} & 61.1 \std{2.6} & 5.4 \std{0.5} & \textbf{46.4} \std{0.2} & \textbf{56.1} \std{0.2} \\
& Sep-Full-Attention $\star$ & \textbf{48.6} \std{2.5} & 28.5 \std{1.6} & \textbf{1.9} \std{0.0} & 58.1 \std{6.9} & 5.1 \std{0.2} & \textbf{46.4} \std{0.2} & 55.4 \std{0.0} \\
& Sep-to-Sep-and-Last $\star$ & 42.4 \std{1.2} & 30.1 \std{0.7} & \textbf{1.9} \std{0.0} & 60.4 \std{1.3} & 6.4 \std{0.6} & 46.2 \std{0.2} & 55.4 \std{0.1} \\
& Sep-to-Last $\star$ & 40.7 \std{1.2} & 24.1 \std{3.3} & 1.8 \std{0.0} & \textbf{62.0} \std{1.9} & 5.5 \std{0.5} & 46.0 \std{0.0} & 55.5 \std{0.2} \\
\midrule

\multirow{7}{*}{\rotatebox{90}{Fine-tuning}}
& \color{gray} Transformer & 58.3 \std{0.1} & 60.6 \std{1.8} & 1.8 \std{0.2} & 91.9 \std{0.8} & 51.4 \std{3.0} & 51.3 \std{0.0} & 66.4 \std{0.3} \\
& ~~$+$ Separators $\star$ & 58.2 \std{0.6} & 61.1 \std{2.2} & 1.9 \std{0.1} & 92.4 \std{0.4} & \textbf{53.6} \std{2.3} & 54.8 \std{0.8} & 67.6 \std{0.4} \\
& \color{gray} Relative Attention & \textbf{61.6} \std{0.5} & \textbf{65.0} \std{0.3} & \textbf{2.7} \std{0.3} & 91.8 \std{1.3} & 51.5 \std{1.2} & 55.6 \std{0.7} & 66.2 \std{1.0} \\
& ~~$+$ Separators $\star$ & 59.9 \std{1.0} & 63.4 \std{0.4} & 2.4 \std{0.3} & \textbf{93.0} \std{0.4} & 45.1 \std{2.7} & \textbf{58.5} \std{1.8} & 67.5 \std{0.3} \\
& Sep-Full-Attention $\star$ & 60.4 \std{1.2} & 62.2 \std{1.4} & 2.2 \std{0.3} & 92.4 \std{1.0} & 48.2 \std{0.6} & 57.2 \std{0.8} & \textbf{67.7} \std{0.3} \\
& Sep-to-Sep-and-Last $\star$ & 58.4 \std{0.6} & 63.0 \std{1.1} & 2.2 \std{0.2} & 92.5 \std{0.5} & 50.2 \std{2.2} & 56.5 \std{0.7} & 66.6 \std{0.9} \\
& Sep-to-Last $\star$ & 59.8 \std{0.7} & 60.7 \std{2.0} & 2.6 \std{0.2} & 92.4 \std{0.5} & 51.3 \std{0.9} & 57.9 \std{0.9} & 67.5 \std{0.5} \\
\bottomrule
\end{tabular}
\vspace{-2mm}
\caption{\textbf{RobustFill} zero-shot generalization and few-shot fine-tuning results.}
\label{tab:robustfill}
\end{table}

All models struggled with zero-shot \SwitchConceptOrder. The ordering pattern is likely very obvious to the Transformer, much like how a language model would learn that sentences start with capital letters and end with punctuation. Again we observe that the easier a pattern is to see, the harder zero-shot generalization would be---the model is more likely to overfit to that particular pattern, making it unlikely to deviate from that pattern when needed for generalization.

For several tasks, in particular \SwitchConceptOrder\ in both domains, few-shot fine-tuning is very effective. Even though the model only sees 20 examples from the test distribution, the best model improves from 6.6\% to 53.6\% for RobustFill, or 17.5\% to 66.8\% for SCAN. With fine-tuning, our Decompositional Transformer models exceed 90\% generalization on RobustFill's \ComposeDiffConcepts\ and SCAN's \ComposeNewOp\ and \AddOpFunctionality.

Finally, we note that none of the three variations of attention masks were consistently better than any others. Thus, it is possible
that the best attention mask type could be application-specific. For instance, Sep-to-Last is the best for zero-shot \ComposeDiffConcepts\ in both domains, which can be explained by the observation that a model would perform well on this generalization task if it can effectively ``forget'' what concept was used in previous program parts, and Sep-to-Last is the most sparse attention mask with the least attention to previous parts.

\section{Related Work}

\textbf{Program Synthesis.}\quad 
For surveys on program synthesis
and machine learning for software engineering, see
\citet{THREEPILLARS,ARMANDOCOURSE,RISHABHSURVEY,BIGCODE}.
Much attention has focused on machine learning for programming by example
\citep{Devlin2017Robustfill,bunel2018leveraging,parisotto2017neuro,DREAMCODER}.
Many methods incorporate learning to guide the search over programs, 
such as using learned premise selection \citep{DEEPCODER,SIGNATURES},
syntax-guided search \citep{Yin2017-my,EUPHONY},
bottom-up search \citep{PROBE,shi2020tf},
two-level search \citep{INFERSKETCHES},
per-example search \citep{PEPS},
and execution-guided synthesis methods \citep{GARBAGECOLLECTOR,REPL,ChenLS19,BUSTLE,CROSSBEAM}.
Another class of program synthesis methods are symbolic search methods \citep{ARMANDOCOURSE,RISHABHSURVEY}, such as bottom-up search
and satisfiability solvers. Since purely symbolic search
will not have problems with compositional generalization
(barring failures or timeouts in search),
it is an intriguing question for future work whether learning-based search methods face challenges with compositional 
generalization.
More generally, there is less work on systematic generalization for machine learning for code,
although \citet{IPAGNN} studies length generalization in the context of the learning-to-execute task~\citep{learning-to-execute}.

\textbf{Compositional Generalization Benchmarks.}\quad
Many datasets have been proposed by the NLP community to evaluate understanding of natural language sentences with compositionally novel structures, such as SCAN~\citep{Lake2018Scan}.
These benchmarks are either constructed by synthesizing examples based on a fine-grained schema of generalization patterns like this work~\citep{Bahdanau2019CLOSUREAS,keysers20cfq,Kim2020COGSAC}, or by repartitioning existing datasets with \textit{i.i.d.}~samples into splits with disjoint compositional structures~\citep{finegan18improving,shaw-etal-2021-compositional}.
Our dataset is related to the COGS benchmark~\citep{Kim2020COGSAC}, which defines a taxonomy of compositional structures in English syntax for natural language understanding.
While many generalization concepts are similar to those proposed in \autoref{sec:generalization} (e.g., extend operation functionality), we focus on measuring and modeling compositional generalization of computer programs under task specifications in both natural language and I/O examples.


\textbf{Improving Compositional Generalization.}\quad
A large body of work develops specialized neural architectures with improved generalization performance~\citep{russin19separate,li19primitive,liu20analytical,chen20neuralsymbolic,herzig20spanparsing}, but is typically limited to specific domains and tasks.
More generalized approaches have been proposed, such as meta-learning~\citep{lake19compositionalmetalearning,wang2020meta,Conklin2021MetaLearningTC}, data augmentation~\citep{andreas20goodenough,oren-etal-2021-finding,Akyrek2021LearningTR,Wang2021LearningTS,qiu2021improving}, and improving the representation of programs~\citep{Furrer2020CompositionalGI,Herzig2021UnlockingCG}.
Related to our work, recent studies attempt to regularize the attention distribution over source tokens to improve generalization of language understanding models~\citep{oren20improvecompositional,yin2021compositional}, which encourage the model to attend to the aligned concept in the input (e.g., \nl{right}) when predicting a function (e.g., \redT{Right}).
Instead of relying on such alignment information between the source and targets to regularize cross-attention distributions, we mask the self-attention scores in the decoder to capture the compositionality of programs, which is applicable to domains like RobustFill where the source-target alignments are not clearly defined.

\section{Conclusion}

We argue that compositional generalization is particularly
important for neural program synthesis, for two reasons. 
On the practical side, we would like synthesizers to be
able to length generalize, generalize to novel combinations of concepts,
and so on.
On the conceptual side, measuring compositional generalization might give us
insight into what problem-solving strategies are learned by neural program synthesizers. 
To that end, we propose a suite of generalization tasks, which measure different
types of compositional generalization that are desirable for program synthesis.
These tasks can be applied to different synthesis domains
to produce a set of benchmarks; we have introduced benchmarks for string
manipulation programs and a simple navigation domain.
We show that these benchmarks are particularly difficult for current 
sequence to sequence models, and we present some early results on modifications
to the Transformer attention mechanism to encourage better generalization.
Future work could explore whether large pre-trained Transformers
also have difficulty with these benchmarks, as well as further methods
for improving compositional generalization.

\bibliography{paper}

\begin{thebibliography}{63}
\providecommand{\natexlab}[1]{#1}
\providecommand{\url}[1]{\texttt{#1}}
\expandafter\ifx\csname urlstyle\endcsname\relax
  \providecommand{\doi}[1]{doi: #1}\else
  \providecommand{\doi}{doi: \begingroup \urlstyle{rm}\Url}\fi

\bibitem[Aky{\"u}rek et~al.(2021)Aky{\"u}rek, Akyurek, and
  Andreas]{Akyrek2021LearningTR}
Ekin Aky{\"u}rek, Afra~Feyza Akyurek, and Jacob Andreas.
\newblock Learning to recombine and resample data for compositional
  generalization.
\newblock \emph{ArXiv}, abs/2010.03706, 2021.

\bibitem[Allamanis et~al.(2018)Allamanis, Barr, Devanbu, and Sutton]{BIGCODE}
Miltiadis Allamanis, Earl~T Barr, Premkumar Devanbu, and Charles Sutton.
\newblock A survey of machine learning for big code and naturalness.
\newblock \emph{ACM Computing Surveys (CSUR)}, 51\penalty0 (4):\penalty0 81,
  2018.

\bibitem[Andreas(2020)]{andreas20goodenough}
Jacob Andreas.
\newblock Good-enough compositional data augmentation.
\newblock In \emph{Proceedings of ACL}, 2020.

\bibitem[Austin et~al.(2021)Austin, Odena, Nye, Bosma, Michalewski, Dohan,
  Jiang, Cai, Terry, Le, and Sutton]{Austin2021-fu}
Jacob Austin, Augustus Odena, Maxwell Nye, Maarten Bosma, Henryk Michalewski,
  David Dohan, Ellen Jiang, Carrie Cai, Michael Terry, Quoc Le, and Charles
  Sutton.
\newblock Program synthesis with large language models.
\newblock August 2021.

\bibitem[Bahdanau et~al.(2016)Bahdanau, Cho, and Bengio]{Attention}
Dzmitry Bahdanau, Kyunghyun Cho, and Yoshua Bengio.
\newblock Neural machine translation by jointly learning to align and
  translate.
\newblock In \emph{International Conference on Learning Representations
  (ICLR)}, 2016.

\bibitem[Bahdanau et~al.(2019)Bahdanau, de~Vries, O'Donnell, Murty, Beaudoin,
  Bengio, and Courville]{Bahdanau2019CLOSUREAS}
Dzmitry Bahdanau, Harm de~Vries, Timothy~J. O'Donnell, Shikhar Murty, Philippe
  Beaudoin, Yoshua Bengio, and Aaron~C. Courville.
\newblock Closure: Assessing systematic generalization of clevr models.
\newblock \emph{ArXiv}, abs/1912.05783, 2019.

\bibitem[Balog et~al.(2017)Balog, Gaunt, Brockschmidt, Nowozin, and
  Tarlow]{DEEPCODER}
Matej Balog, Alexander~L Gaunt, Marc Brockschmidt, Sebastian Nowozin, and
  Daniel Tarlow.
\newblock {DeepCoder: Learning to write programs}.
\newblock In \emph{International Conference on Learning Representations
  ({ICLR})}, 2017.

\bibitem[Barke et~al.(2020)Barke, Peleg, and Polikarpova]{PROBE}
Shraddha Barke, Hila Peleg, and Nadia Polikarpova.
\newblock {Just-in-Time} learning for {Bottom-Up} enumerative synthesis.
\newblock In \emph{Object-oriented Programming, Systems, Languages, and
  Applications ({{OOPSLA}})}, 2020.

\bibitem[Bieber et~al.(2020)Bieber, Sutton, Larochelle, and Tarlow]{IPAGNN}
David Bieber, Charles Sutton, Hugo Larochelle, and Daniel Tarlow.
\newblock Learning to execute programs with instruction pointer attention graph
  neural networks.
\newblock In \emph{Advances in Neural Information Processing Systems
  ({NeurIPS})}, 2020.

\bibitem[Bunel et~al.(2018)Bunel, Hausknecht, Devlin, Singh, and
  Kohli]{bunel2018leveraging}
Rudy Bunel, Matthew Hausknecht, Jacob Devlin, Rishabh Singh, and Pushmeet
  Kohli.
\newblock Leveraging grammar and reinforcement learning for neural program
  synthesis.
\newblock In \emph{International Conference on Learning Representations}, 2018.

\bibitem[Chen et~al.(2021)Chen, Tworek, Jun, Yuan, Pinto, Kaplan, Edwards,
  Burda, Joseph, Brockman, et~al.]{chen2021evaluating}
Mark Chen, Jerry Tworek, Heewoo Jun, Qiming Yuan, Henrique Ponde de~Oliveira
  Pinto, Jared Kaplan, Harri Edwards, Yuri Burda, Nicholas Joseph, Greg
  Brockman, et~al.
\newblock Evaluating large language models trained on code.
\newblock \emph{arXiv preprint arXiv:2107.03374}, 2021.

\bibitem[Chen et~al.(2019)Chen, Liu, and Song]{ChenLS19}
Xinyun Chen, Chang Liu, and Dawn Song.
\newblock Execution-guided neural program synthesis.
\newblock In \emph{International Conference on Learning Representations
  (ICLR)}, 2019.

\bibitem[Chen et~al.(2020)Chen, Liang, Yu, Song, and
  Zhou]{chen20neuralsymbolic}
Xinyun Chen, Chen Liang, Adams~Wei Yu, D.~Song, and Denny Zhou.
\newblock Compositional generalization via neural-symbolic stack machines.
\newblock In \emph{Proceedings of NeurIPS}, 2020.

\bibitem[Chomsky \& Lightfoot(2002)Chomsky and Lightfoot]{chomsky2002syntactic}
N.~Chomsky and D.W. Lightfoot.
\newblock \emph{Syntactic Structures}.
\newblock De Gruyter Reference Global. Mouton de Gruyter, 2002.
\newblock ISBN 9783110172799.
\newblock URL \url{https://books.google.com/books?id=a6a\_b-CXYAkC}.

\bibitem[Conklin et~al.(2021)Conklin, Wang, Smith, and
  Titov]{Conklin2021MetaLearningTC}
Henry Conklin, Bailin Wang, Kenny Smith, and Ivan Titov.
\newblock Meta-learning to compositionally generalize.
\newblock \emph{ArXiv}, abs/2106.04252, 2021.

\bibitem[Csord{\'a}s et~al.(2021)Csord{\'a}s, Irie, and
  Schmidhuber]{Csordas2021-ey}
R{\'o}bert Csord{\'a}s, Kazuki Irie, and J{\"u}rgen Schmidhuber.
\newblock The devil is in the detail: Simple tricks improve systematic
  generalization of transformers.
\newblock August 2021.

\bibitem[Devlin et~al.(2017)Devlin, Uesato, Bhupatiraju, Singh, Mohamed, and
  Kohli]{Devlin2017Robustfill}
Jacob Devlin, Jonathan Uesato, Surya Bhupatiraju, Rishabh Singh, Abdel{-}rahman
  Mohamed, and Pushmeet Kohli.
\newblock {RobustFill}: Neural program learning under noisy {I/O}.
\newblock \emph{ICML}, 2017.

\bibitem[Ellis et~al.(2019)Ellis, Nye, Pu, Sosa, Tenenbaum, and
  Solar{-}Lezama]{REPL}
Kevin Ellis, Maxwell~I. Nye, Yewen Pu, Felix Sosa, Josh Tenenbaum, and Armando
  Solar{-}Lezama.
\newblock Write, execute, assess: Program synthesis with a {REPL}.
\newblock In \emph{Neural Information Processing Systems (NeurIPS)}, 2019.

\bibitem[Ellis et~al.(2020)Ellis, Wong, Nye, Sable-Meyer, Cary, Morales,
  Hewitt, Solar-Lezama, and Tenenbaum]{DREAMCODER}
Kevin Ellis, Catherine Wong, Maxwell Nye, Mathias Sable-Meyer, Luc Cary, Lucas
  Morales, Luke Hewitt, Armando Solar-Lezama, and Joshua~B. Tenenbaum.
\newblock Dreamcoder: Growing generalizable, interpretable knowledge with
  wake-sleep bayesian program learning.
\newblock \emph{CoRR}, abs/2006.08381, 2020.
\newblock URL \url{https://arxiv.org/abs/2006.08381}.

\bibitem[Finegan-Dollak et~al.(2018)Finegan-Dollak, Kummerfeld, Zhang,
  Ramanathan, Sadasivam, Zhang, and Radev]{finegan18improving}
Catherine Finegan-Dollak, Jonathan~K. Kummerfeld, Li~Zhang, Karthik Ramanathan,
  Sesh Sadasivam, Rui Zhang, and Dragomir Radev.
\newblock Improving text-to-{SQL} evaluation methodology.
\newblock In \emph{Proceedings of ACL}, 2018.

\bibitem[Furrer et~al.(2020)Furrer, van Zee, Scales, and
  Scharli]{Furrer2020CompositionalGI}
Daniel Furrer, Marc van Zee, Nathan Scales, and Nathanael Scharli.
\newblock Compositional generalization in semantic parsing: Pre-training vs.
  specialized architectures.
\newblock \emph{ArXiv}, abs/2007.08970, 2020.

\bibitem[Gottschlich et~al.(2018)Gottschlich, Solar-Lezama, Tatbul, Carbin,
  Rinard, Barzilay, Amarasinghe, Tenenbaum, and Mattson]{THREEPILLARS}
Justin Gottschlich, Armando Solar-Lezama, Nesime Tatbul, Michael Carbin, Martin
  Rinard, Regina Barzilay, Saman Amarasinghe, Joshua~B Tenenbaum, and Tim
  Mattson.
\newblock The three pillars of machine programming.
\newblock In \emph{ACM SIGPLAN International Workshop on Machine Learning and
  Programming Languages}, pp.\  69--80. ACM, 2018.

\bibitem[Gu et~al.(2021)Gu, Kase, Vanni, Sadler, Liang, Yan, and
  Su]{gu2021beyond}
Yu~Gu, Sue Kase, Michelle Vanni, Brian Sadler, Percy Liang, Xifeng Yan, and
  Yu~Su.
\newblock Beyond iid: three levels of generalization for question answering on
  knowledge bases.
\newblock In \emph{Proceedings of the Web Conference 2021}, pp.\  3477--3488,
  2021.

\bibitem[Gulwani et~al.(2017{\natexlab{a}})Gulwani, Polozov, and
  Singh]{gulwani2017program}
S.~Gulwani, O.~Polozov, and R.~Singh.
\newblock \emph{Program Synthesis}.
\newblock Foundations and Trends(r) in Programming Languages Series. Now
  Publishers, 2017{\natexlab{a}}.
\newblock ISBN 9781680832921.
\newblock URL \url{https://books.google.com/books?id=mK5ctAEACAAJ}.

\bibitem[Gulwani(2011)]{FLASHFILL}
Sumit Gulwani.
\newblock Automating string processing in spreadsheets using input-output
  examples.
\newblock In \emph{PoPL'11, January 26-28, 2011, Austin, Texas, USA}, 2011.

\bibitem[Gulwani et~al.(2017{\natexlab{b}})Gulwani, Polozov, Singh,
  et~al.]{RISHABHSURVEY}
Sumit Gulwani, Oleksandr Polozov, Rishabh Singh, et~al.
\newblock Program synthesis.
\newblock \emph{Foundations and Trends{\textregistered} in Programming
  Languages}, 4\penalty0 (1-2):\penalty0 1--119, 2017{\natexlab{b}}.

\bibitem[Herzig \& Berant(2020)Herzig and Berant]{herzig20spanparsing}
Jonathan Herzig and Jonathan Berant.
\newblock Span-based semantic parsing for compositional generalization.
\newblock In \emph{Proceedings of EMNLP}, 2020.

\bibitem[Herzig et~al.(2021)Herzig, Shaw, Chang, Guu, Pasupat, and
  Zhang]{Herzig2021UnlockingCG}
Jonathan Herzig, Peter Shaw, Ming-Wei Chang, Kelvin Guu, Panupong Pasupat, and
  Yuan Zhang.
\newblock Unlocking compositional generalization in pre-trained models using
  intermediate representations.
\newblock \emph{ArXiv}, abs/2104.07478, 2021.

\bibitem[Keysers et~al.(2020)Keysers, Sch{\"a}rli, Scales, Buisman, Furrer,
  Kashubin, Momchev, Sinopalnikov, Stafiniak, Tihon, Tsarkov, Wang, van Zee,
  and Bousquet]{keysers20cfq}
Daniel Keysers, Nathanael Sch{\"a}rli, Nathan Scales, H.~Buisman, Daniel
  Furrer, Sergii Kashubin, Nikola Momchev, Danila Sinopalnikov, Lukasz
  Stafiniak, Tibor Tihon, D.~Tsarkov, Xiao Wang, Marc van Zee, and O.~Bousquet.
\newblock Measuring compositional generalization: A comprehensive method on
  realistic data.
\newblock In \emph{Proceedings of ICLR}, 2020.

\bibitem[Kim \& Linzen(2020)Kim and Linzen]{Kim2020COGSAC}
Najoung Kim and Tal Linzen.
\newblock Cogs: A compositional generalization challenge based on semantic
  interpretation.
\newblock \emph{ArXiv}, abs/2010.05465, 2020.

\bibitem[Lake(2019)]{lake19compositionalmetalearning}
Brenden~M Lake.
\newblock Compositional generalization through meta sequence-to-sequence
  learning.
\newblock In \emph{Proceedings of NeurIPS}, 2019.

\bibitem[Lake \& Baroni(2018)Lake and Baroni]{Lake2018Scan}
Brenden~M. Lake and Marco Baroni.
\newblock Generalization without systematicity: On the compositional skills of
  sequence-to-sequence recurrent networks.
\newblock \emph{ICML}, 2018.

\bibitem[Lee et~al.(2018)Lee, Heo, Alur, and Naik]{EUPHONY}
Woosuk Lee, Kihong Heo, Rajeev Alur, and Mayur Naik.
\newblock Accelerating search-based program synthesis using learned
  probabilistic models.
\newblock In \emph{Conference on Programming Language Design and Implementation
  (PLDI)}, pp.\  436--449, June 2018.

\bibitem[Li et~al.(2019)Li, Zhao, Wang, and Hestness]{li19primitive}
Yuanpeng Li, Liang Zhao, JianYu Wang, and Joel Hestness.
\newblock Compositional generalization for primitive substitutions.
\newblock In \emph{Proceedings of EMNLP/IJCNLP}, 2019.

\bibitem[Li et~al.(2022)Li, Choi, Chung, Kushman, Schrittwieser, Leblond,
  Eccles, Keeling, Gimeno, Dal~Lago, Hubert, Choy, De~Masson~D'autume,
  Babuschkin, Chen, Huang, Welbl, Gowal, Cherepanov, Molloy, Mankowitz, Robson,
  Kohli, De~Freitas, Kavukcuoglu, and Vinyals]{ALPHACODE}
Yujia Li, David Choi, Junyoung Chung, Nate Kushman, Julian Schrittwieser,
  R{\'e}mi Leblond, Tom Eccles, James Keeling, Felix Gimeno, Agustin Dal~Lago,
  Thomas Hubert, Peter Choy, Cyprien De~Masson~D'autume, Igor Babuschkin,
  Xinyun Chen, Po-Sen Huang, Johannes Welbl, Sven Gowal, Alexey Cherepanov,
  James Molloy, Daniel~J Mankowitz, Esme~Sutherland Robson, Pushmeet Kohli,
  Nando De~Freitas, Koray Kavukcuoglu, and Oriol Vinyals.
\newblock {Competition-Level} code generation with {AlphaCode}.
\newblock
  \url{https://storage.googleapis.com/deepmind-media/AlphaCode/competition_level_code_generation_with_alphacode.pdf},
  2022.
\newblock Accessed: 2022-2-26.

\bibitem[Liu et~al.(2020)Liu, An, Lou, Chen, Lin, Gao, Zhou, Zheng, and
  Zhang]{liu20analytical}
Qian Liu, Shengnan An, Jianguang Lou, B.~Chen, Zeqi Lin, Yan Gao, Bin Zhou,
  Nanning Zheng, and Dongmei Zhang.
\newblock Compositional generalization by learning analytical expressions.
\newblock In \emph{Proceedings of NeurIPS}, 2020.

\bibitem[Marcus(2001)]{Marcus2001TheAM}
Gary~F. Marcus.
\newblock The algebraic mind: Integrating connectionism and cognitive science.
\newblock 2001.

\bibitem[Nye et~al.(2019)Nye, Hewitt, Tenenbaum, and
  Solar{-}Lezama]{INFERSKETCHES}
Maxwell~I. Nye, Luke~B. Hewitt, Joshua~B. Tenenbaum, and Armando
  Solar{-}Lezama.
\newblock Learning to infer program sketches.
\newblock In \emph{International Conference on Machine Learning {(ICML)}}, pp.\
   4861--4870, 2019.
\newblock URL \url{http://proceedings.mlr.press/v97/nye19a.html}.

\bibitem[Odena \& Sutton(2020)Odena and Sutton]{SIGNATURES}
Augustus Odena and Charles Sutton.
\newblock Learning to represent programs with property signatures.
\newblock In \emph{International Conference on Learning Representations
  {(ICLR)}}, 2020.

\bibitem[Odena et~al.(2020)Odena, Shi, Bieber, Singh, Sutton, and Dai]{BUSTLE}
Augustus Odena, Kensen Shi, David Bieber, Rishabh Singh, Charles Sutton, and
  Hanjun Dai.
\newblock {BUSTLE}: {Bottom-Up} program synthesis through learning-guided
  exploration.
\newblock In \emph{International Conference on Learning Representations
  ({ICLR})}, September 2020.

\bibitem[Oren et~al.(2020)Oren, Herzig, Gupta, Gardner, and
  Berant]{oren20improvecompositional}
Inbar Oren, Jonathan Herzig, Nitish Gupta, Matt Gardner, and Jonathan Berant.
\newblock Improving compositional generalization in semantic parsing.
\newblock In \emph{Proceedings of EMNLP{-}Findings}, 2020.

\bibitem[Oren et~al.(2021)Oren, Herzig, and Berant]{oren-etal-2021-finding}
Inbar Oren, Jonathan Herzig, and Jonathan Berant.
\newblock Finding needles in a haystack: Sampling structurally-diverse training
  sets from synthetic data for compositional generalization.
\newblock In \emph{Proceedings of EMNLP}, 2021.

\bibitem[Parisotto et~al.(2017)Parisotto, Mohamed, Singh, Li, Zhou, and
  Kohli]{parisotto2017neuro}
Emilio Parisotto, Abdel{-}rahman Mohamed, Rishabh Singh, Lihong Li, Dengyong
  Zhou, and Pushmeet Kohli.
\newblock Neuro-symbolic program synthesis.
\newblock In \emph{International Conference on Learning Representations
  (ICLR)}, 2017.

\bibitem[Parlante(2022)]{cs106a}
Nick Parlante.
\newblock {CS106A}: Programming methodologies.
\newblock Course at Stanford University, 2022.

\bibitem[Qiu et~al.(2021)Qiu, Shaw, Pasupat, Nowak, Linzen, Sha, and
  Toutanova]{qiu2021improving}
Linlu Qiu, Peter Shaw, Panupong Pasupat, Pawe{\l}~Krzysztof Nowak, Tal Linzen,
  Fei Sha, and Kristina Toutanova.
\newblock Improving compositional generalization with latent structure and data
  augmentation.
\newblock \emph{arXiv preprint arXiv:2112.07610}, 2021.

\bibitem[Russin et~al.(2019)Russin, Jo, O'Reilly, and Bengio]{russin19separate}
Jake Russin, Jason Jo, R.~O'Reilly, and Yoshua Bengio.
\newblock Compositional generalization in a deep seq2seq model by separating
  syntax and semantics.
\newblock \emph{ArXiv}, abs/1904.09708, 2019.

\bibitem[Shaw et~al.(2018)Shaw, Uszkoreit, and Vaswani]{shaw2018relative}
Peter Shaw, Jakob Uszkoreit, and Ashish Vaswani.
\newblock Self-attention with relative position representations.
\newblock In \emph{North American Chapter of the Association for Computational
  Linguistics {(NAACL)}}, 2018.

\bibitem[Shaw et~al.(2021)Shaw, Chang, Pasupat, and
  Toutanova]{shaw-etal-2021-compositional}
Peter Shaw, Ming-Wei Chang, Panupong Pasupat, and Kristina Toutanova.
\newblock Compositional generalization and natural language variation: Can a
  semantic parsing approach handle both?
\newblock In \emph{Proceedings of ACL}, 2021.

\bibitem[Shi et~al.(2019)Shi, Steinhardt, and Liang]{shi2019frangel}
Kensen Shi, Jacob Steinhardt, and Percy Liang.
\newblock {FrAngel}: Component-based synthesis with control structures.
\newblock \emph{Proceedings of the ACM on Programming Languages}, 3\penalty0
  (POPL):\penalty0 1--29, 2019.

\bibitem[Shi et~al.(2020)Shi, Bieber, and Singh]{shi2020tf}
Kensen Shi, David Bieber, and Rishabh Singh.
\newblock {TF-Coder}: Program synthesis for tensor manipulations.
\newblock \emph{arXiv preprint arXiv:2003.09040}, 2020.

\bibitem[Shi et~al.(2022)Shi, Dai, Ellis, and Sutton]{CROSSBEAM}
Kensen Shi, Hanjun Dai, Kevin Ellis, and Charles Sutton.
\newblock {CrossBeam}: Learning to search in bottom-up program synthesis.
\newblock In \emph{International Conference on Learning Representations
  ({ICLR})}, 2022.

\bibitem[Shrivastava et~al.()Shrivastava, Larochelle, and Tarlow]{PEPS}
Disha Shrivastava, Hugo Larochelle, and Daniel Tarlow.
\newblock Learning to combine per-example solutions for neural program
  synthesis.
\newblock In \emph{Advances in Neural Information Processing Systems}.

\bibitem[Solar-Lezama(2018)]{ARMANDOCOURSE}
Armando Solar-Lezama.
\newblock Introduction to program synthesis.
\newblock \url{https://people.csail.mit.edu/asolar/SynthesisCourse/TOC.htma},
  2018.
\newblock Accessed: 2018-09-17.

\bibitem[Solar{-}Lezama et~al.(2006)Solar{-}Lezama, Tancau, Bod{\'{\i}}k,
  Seshia, and Saraswat]{SKETCH}
Armando Solar{-}Lezama, Liviu Tancau, Rastislav Bod{\'{\i}}k, Sanjit~A. Seshia,
  and Vijay~A. Saraswat.
\newblock Combinatorial sketching for finite programs.
\newblock In \emph{Conference on Architectural Support for Programming
  Languages and Operating Systems, {ASPLOS} 2006, San Jose, CA, USA, October
  21-25, 2006}, pp.\  404--415. {ACM}, 2006.

\bibitem[Talmor \& Berant(2018)Talmor and Berant]{talmor2018web}
Alon Talmor and Jonathan Berant.
\newblock The web as a knowledge-base for answering complex questions.
\newblock In \emph{Proceedings of the 2018 Conference of the North American
  Chapter of the Association for Computational Linguistics: Human Language
  Technologies, Volume 1 (Long Papers)}, pp.\  641--651, 2018.

\bibitem[Torlak \& Bod{\'{\i}}k(2013)Torlak and Bod{\'{\i}}k]{ROSETTE}
Emina Torlak and Rastislav Bod{\'{\i}}k.
\newblock Growing solver-aided languages with rosette.
\newblock In Antony~L. Hosking, Patrick~Th. Eugster, and Robert Hirschfeld
  (eds.), \emph{{ACM} Symposium on New Ideas in Programming and Reflections on
  Software, Onward! 2013, part of {SPLASH} '13, Indianapolis, IN, USA, October
  26-31, 2013}, pp.\  135--152. {ACM}, 2013.
\newblock \doi{10.1145/2509578.2509586}.
\newblock URL \url{https://doi.org/10.1145/2509578.2509586}.

\bibitem[Vaswani et~al.(2017)Vaswani, Shazeer, Parmar, Uszkoreit, Jones, Gomez,
  Kaiser, and Polosukhin]{Transformer}
Ashish Vaswani, Noam Shazeer, Niki Parmar, Jakob Uszkoreit, Llion Jones,
  Aidan~N. Gomez, Lukasz Kaiser, and Illia Polosukhin.
\newblock Attention is all you need.
\newblock In \emph{Neural Information Processing Systems (NeurIPS)}, 2017.

\bibitem[Wang et~al.(2020)Wang, Lapata, and Titov]{wang2020meta}
Bailin Wang, Mirella Lapata, and Ivan Titov.
\newblock Meta-learning for domain generalization in semantic parsing.
\newblock \emph{arXiv:2010.11988}, 2020.

\bibitem[Wang et~al.(2021)Wang, Yin, Lin, and Xiong]{Wang2021LearningTS}
Bailin Wang, Wenpeng Yin, Xi~Victoria Lin, and Caiming Xiong.
\newblock Learning to synthesize data for semantic parsing.
\newblock \emph{ArXiv}, abs/2104.05827, 2021.

\bibitem[Yin \& Neubig(2017)Yin and Neubig]{Yin2017-my}
Pengcheng Yin and Graham Neubig.
\newblock A syntactic neural model for {General-Purpose} code generation.
\newblock In \emph{Assocation for Computational Linguistics {(ACL)}}, 2017.

\bibitem[Yin et~al.(2021)Yin, Fang, Neubig, Pauls, Platanios, Su, Thomson, and
  Andreas]{yin2021compositional}
Pengcheng Yin, Hao Fang, Graham Neubig, Adam Pauls, Emmanouil~Antonios
  Platanios, Yu~Su, Sam Thomson, and Jacob Andreas.
\newblock Compositional generalization for neural semantic parsing via
  span-level supervised attention.
\newblock In \emph{2021 Conference of the North American Chapter of the
  Association for Computational Linguistics: Human Language Technologies},
  2021.

\bibitem[Zaremba \& Sutskever(2014)Zaremba and Sutskever]{learning-to-execute}
Wojciech Zaremba and Ilya Sutskever.
\newblock Learning to execute, 2014.

\bibitem[Zohar \& Wolf(2018)Zohar and Wolf]{GARBAGECOLLECTOR}
Amit Zohar and Lior Wolf.
\newblock Automatic program synthesis of long programs with a learned garbage
  collector.
\newblock In \emph{Neural Information Processing Systems (NeurIPS)}, 2018.

\end{thebibliography}
\bibliographystyle{iclr2022_conference}

\clearpage

\appendix
\appendixpage

\section{SCAN and RobustFill DSLs}
\label{app:dsls}

\autoref{fig:scan_dsl} contains the DSL for SCAN translation tasks, and
\autoref{fig:robustfill_dsl} contains the DSL for our RobustFill programs.

\begin{figure*}[ht]
\small
\begin{alignat*}{2}
\mbox{Command } C\quad &:= &\quad& C_1 \ \nl{\text{after}} \ C_2 \logicalOR D \\
D\quad &:= && D_1 \ \nl{\text{and}} \ D_2 \logicalOR P \\
\mbox{Part } P\quad &:= && Q \logicalOR Q \ \nl{\text{twice}} \logicalOR Q \ \nl{\text{thrice}} \\
Q\quad &:= && l \logicalOR r \logicalOR a \\
\mbox{Left-concept } l\quad &:= && v \ \nl{\text{left}} \logicalOR v \ \nl{\text{opposite left}} \logicalOR v \ \nl{ \text{around left}} \\
\mbox{Right-concept } r\quad &:= && v \ \nl{\text{right}} \logicalOR v \ \nl{\text{opposite right}} \logicalOR v \ \nl{ \text{around right}} \\
\mbox{Verb } v\quad &:= && \nl{\text{turn}} \logicalOR a \\
\mbox{Action } a\quad &:= && \nl{\text{walk}} \logicalOR \nl{\text{look}} \logicalOR \nl{\text{run}} \logicalOR \nl{\text{jump}} \\
\\[-1.0em]\hline\\[-1.0em]
C_1 \ \nl{\text{after}} \ C_2  &\rightarrow && C_2 \ C_1 \\
D_1 \ \nl{\text{and}} \ D_2 &\rightarrow && D_1 \ D_2 \\
Q \ \nl{\text{twice}} &\rightarrow && Q \ Q \\
Q \ \nl{\text{thrice}} &\rightarrow && Q \ Q \ Q \\
\nl{\text{turn left}} &\rightarrow && \redT{LTURN} \\
\nl{\text{turn right}} &\rightarrow && \redT{RTURN} \\
a \ \nl{\text{left}} &\rightarrow && \redT{LTURN} \ a \\
a \ \nl{\text{right}} &\rightarrow && \redT{RTURN} \ a \\
\nl{\text{turn opposite left}} &\rightarrow && \redT{LTURN LTURN} \\
\nl{\text{turn opposite right}} &\rightarrow && \redT{RTURN RTURN} \\
a \ \nl{\text{opposite left}} &\rightarrow && \redT{LTURN LTURN} \ a \\
a \ \nl{\text{opposite right}} &\rightarrow && \redT{RTURN RTURN} \ a \\
\nl{\text{turn around left}} &\rightarrow && \redT{LTURN LTURN LTURN LTURN} \\
\nl{\text{turn around right}} &\rightarrow && \redT{RTURN RTURN RTURN RTURN} \\
a \ \nl{\text{around left}} &\rightarrow &&  \redT{LTURN} \ a \ \redT{LTURN} \ a \ \redT{LTURN} \ a \ \redT{LTURN} \ a \\
a \ \nl{\text{around right}} &\rightarrow && \redT{RTURN} \ a \ \redT{RTURN} \ a \ \redT{RTURN} \ a \ \redT{RTURN} \ a \\
\nl{\text{walk}} &\rightarrow && \redT{WALK} \\
\nl{\text{look}} &\rightarrow && \redT{LOOK} \\
\nl{\text{run}} &\rightarrow && \redT{RUN} \\
\nl{\text{jump}} &\rightarrow && \redT{JUMP} \\
\end{alignat*}
    \caption{The DSL for SCAN commands (top) with translations (bottom) from language-like commands (e.g., \nl{jump left}) to action sequences (e.g., \redT{LTURN JUMP}). This is generalized from \citet{Lake2018Scan} to allow arbitrarily-many \nl{and} and \nl{after} conjunctions.} 
\label{fig:scan_dsl}
\end{figure*}

\begin{figure*}[ht]
\small
\begin{alignat*}{2}
\mbox{Program } P\quad &:= &\quad& \T{Concat}(e_1, e_2, \hdots) \\
\mbox{Expression } e\quad &:= && s \logicalOR m \logicalOR o \logicalOR \T{ConstStr}(c) \\
\mbox{Compose } o\quad &:= && m_1(m_2) \logicalOR m(s) \\
\mbox{Substring } s\quad &:= && \T{SubStr}(k_1, k_2) \logicalOR 
\T{GetSpan}(r_1, i_1, b_1, r_2, i_2, b_2) \logicalOR \T{GetToken}(t, i) \\
&&& \logicalOR \T{GetUpto}(r) 
\logicalOR \T{GetFrom}(r)  \\
\mbox{Modification } m\quad &:= && \T{ToCase}(a) \logicalOR \T{Replace}(\delta_1, \delta_2) \logicalOR \T{Trim}() \logicalOR \T{GetFirst}(t, i) \logicalOR \T{GetAll}(t) \\
&&&\logicalOR  \T{Substitute}(t, i, c) \logicalOR  \T{SubstituteAll}(t, c) \logicalOR \T{Remove}(t, i) \logicalOR \T{RemoveAll}(t) \\
\mbox{Regex } r\quad &:= && t_1 \logicalOR \hdots \logicalOR t_n \logicalOR \delta_1 \logicalOR \hdots \logicalOR \delta_m \\
\mbox{Type } t\quad &:= &&  \T{NUMBER} \logicalOR \T{WORD} \logicalOR \T{ALPHANUM} \logicalOR \T{ALL\_CAPS} \logicalOR \T{PROP\_CASE} \logicalOR \T{LOWER} \logicalOR \T{DIGIT} \logicalOR \T{CHAR} \\
\mbox{Case } a\quad &:= && \T{PROPER} \logicalOR \T{ALL\_CAPS} \logicalOR \T{LOWER} \\
\mbox{Position } k\quad &:= && -100 \logicalOR -99 \logicalOR \hdots \logicalOR 1 \logicalOR 2 \logicalOR \hdots \logicalOR 100 \\
\mbox{Index } i\quad &:= && -5 \logicalOR -4 \logicalOR \hdots \logicalOR -1 \logicalOR 1 \logicalOR 2 \logicalOR \hdots \logicalOR 5 \\
\mbox{Boundary } b\quad &:= && \T{START} \logicalOR \T{END} \\
\mbox{Delimiter } \delta\quad &:= && \&\,,.?@()[]\%\{\}/:;\$\#"' \\
\mbox{Character } c\quad &:= && A-Z \logicalOR a-z \logicalOR 0-9 \logicalOR \&,.?@\hdots
\end{alignat*}
    \caption{The DSL for string transformation tasks in the RobustFill domain, slightly modified from \citep{Devlin2017Robustfill} to add more functionality.} 
\label{fig:robustfill_dsl}
\end{figure*}

\section{SCAN and RobustFill Generalization Task Details}
\label{app:benchmark_details}

\kensen{technically we should switch the ordering of the two paragraphs, but that also requires significant rewording since the SCAN paragraph just lists differences}

\paragraph{RobustFill Details.}
Unless stated otherwise, all programs described in the following paragraph have length between 2 and 6 (number of program parts). For \ComposeDiffConcepts, we group together all of the substring operations into \emph{substring concepts} and all of the modification operations plus constant strings as \emph{non-substring concepts} (the Compose operation is omitted). We use the same concepts for \SwitchConceptOrder, where training examples use only the substring concept for the first half of the parts and only the non-substring concept for the latter half, and test examples have the ordering reversed.\todo{It might be more clear to use the $A$ and $B$ notation
from the previous section, rather than the "first half of the parts", which is a concept that has not been referred to before.}
\kensen{i was afraid the A/B notation would imply that there are only 2 parts of the program. if there are 6 parts, i want to say that the first 3 are of A concept, and the latter 3 are of B concept. furthermore this is a decision we made that isn't necessarily the case in general}
For \ComposeNewOp, 25\% of training examples are length 1 programs containing only a Compose operation, the remainder of the training examples are length 2-6 programs without the Compose operation, and the test examples are length 2-6 programs that use the Compose operation. For \AddOpFunctionality, all examples are length 1-6 programs, training examples are those where a substring operation is not used within a Compose operation, and test examples are those where a substring operation is used within a Compose operation.

\paragraph{SCAN Details.}
We create compositional generalization tasks for the SCAN domain in largely the same way as for the RobustFill domain, with the following differences. For \ComposeDiffConcepts\ and \SwitchConceptOrder, we use all left-direction phrases as one concept, all right-direction phrases as the other concept, and omit phrases without a direction. For \ComposeNewOp, 10\% of training examples are (exactly) the length 1 \nl{jump} command, the remaining training examples are length 1-6 commands that do not contain \nl{jump}, and the training examples are length 1-6 commands that do contain \nl{jump} (but are not exactly \nl{jump} itself). For \AddOpFunctionality, training commands do not contain \nl{around right}, while test commands do contain \nl{around right}. The setup of the latter two generalization tasks closely mirrors tasks from \citet{Lake2018Scan}.

\section{Model and Training Hyperparameters}
\label{app:hyperparams}

For our models, we used default hyperparameters already existing in the frameworks used in our implementation. In particular, the Transformer has 4 attention heads, 3 layers, a hidden dimension of 512, and an embedding dimension of 256. Relative attention uses 32 different buckets for relative positions, with a maximum distance of 128. We use a dropout rate of 0.1. During training, we use a learning rate schedule consisting of a base learning rate of $1\times10^{-3}$, linear warmup of 16000 steps, and square root decay. During fine-tuning, we use a constant learning rate of $1\times10^{-4}$.

\section{Plots for SCAN and RobustFill Results}
\label{app:plots}

\begin{figure*}[ht]
\centering
\includegraphics[width=\textwidth]{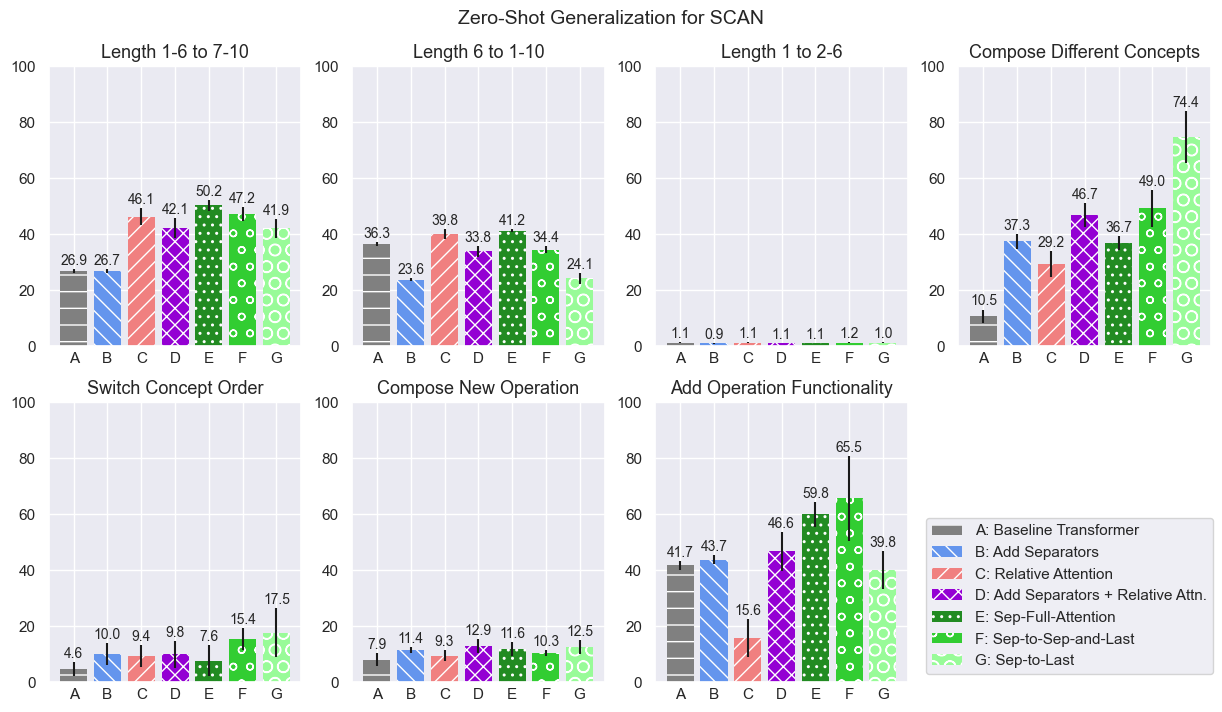}
\vspace{-2em}
\caption{Zero-shot generalization for SCAN, trained for 10,000 steps with a batch size of 128. The bar heights represent the mean accuracy on the test set over 3 different random initializations, and the error bars represent one standard deviation above and below the mean.}
\label{fig:zero-shot-scan}
\end{figure*}

\begin{figure*}[ht]
\centering
\includegraphics[width=\textwidth]{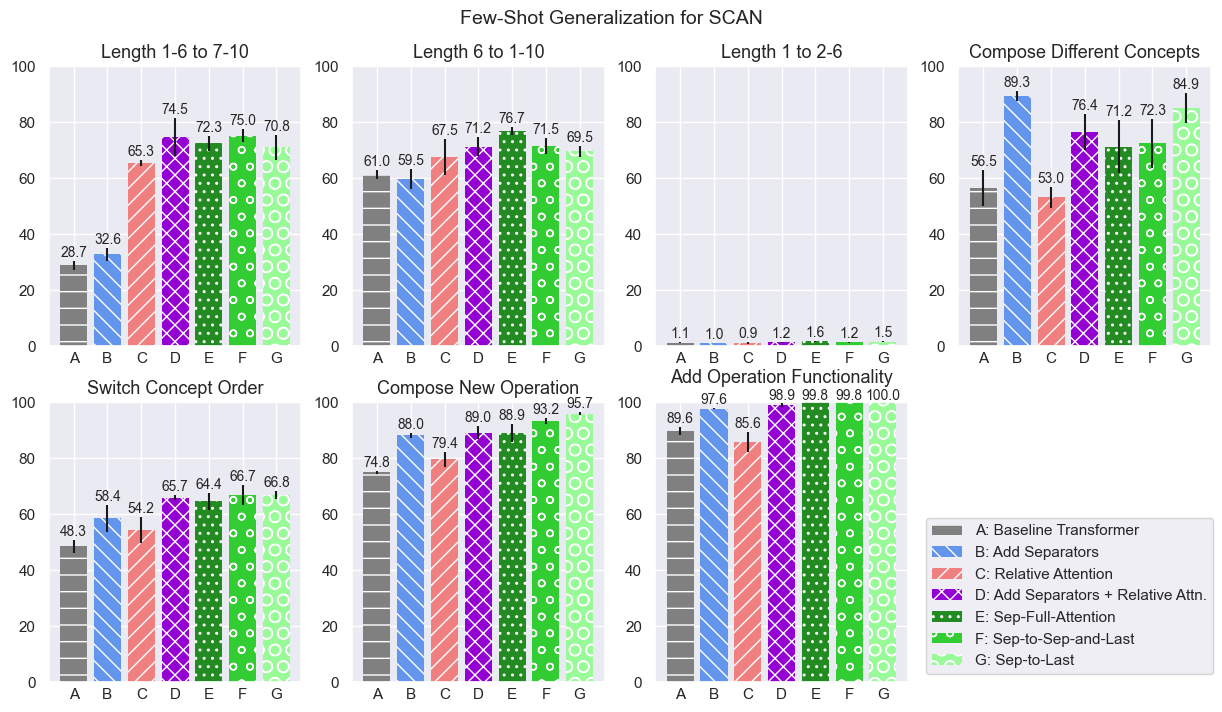}
\vspace{-2em}
\caption{Few-shot generalization for SCAN, fine-tuning with 20 examples from the train distribution and 20 examples from the test distribution, for 30 epochs. As in the zero-shot case, the bars show the mean accuracy on the test set over the 3 random initializations with error bars representing one standard deviation above and below the mean.}
\label{fig:few-shot-scan}
\end{figure*}

\begin{figure*}[ht]
\centering
\includegraphics[width=\textwidth]{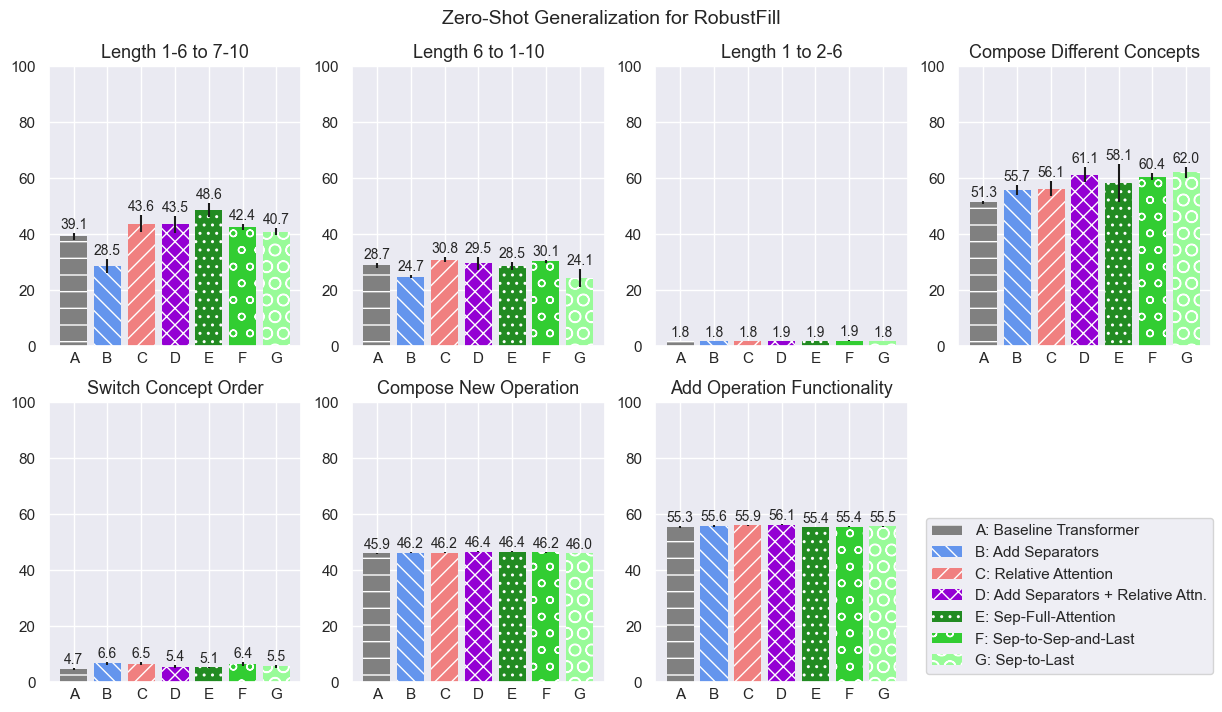}
\vspace{-2em}
\caption{Zero-shot generalization for RobustFill, trained for 1 million steps with a batch size of 128.}
\label{fig:zero-shot-robustfill}
\end{figure*}

\begin{figure*}[ht]
\centering
\includegraphics[width=\textwidth]{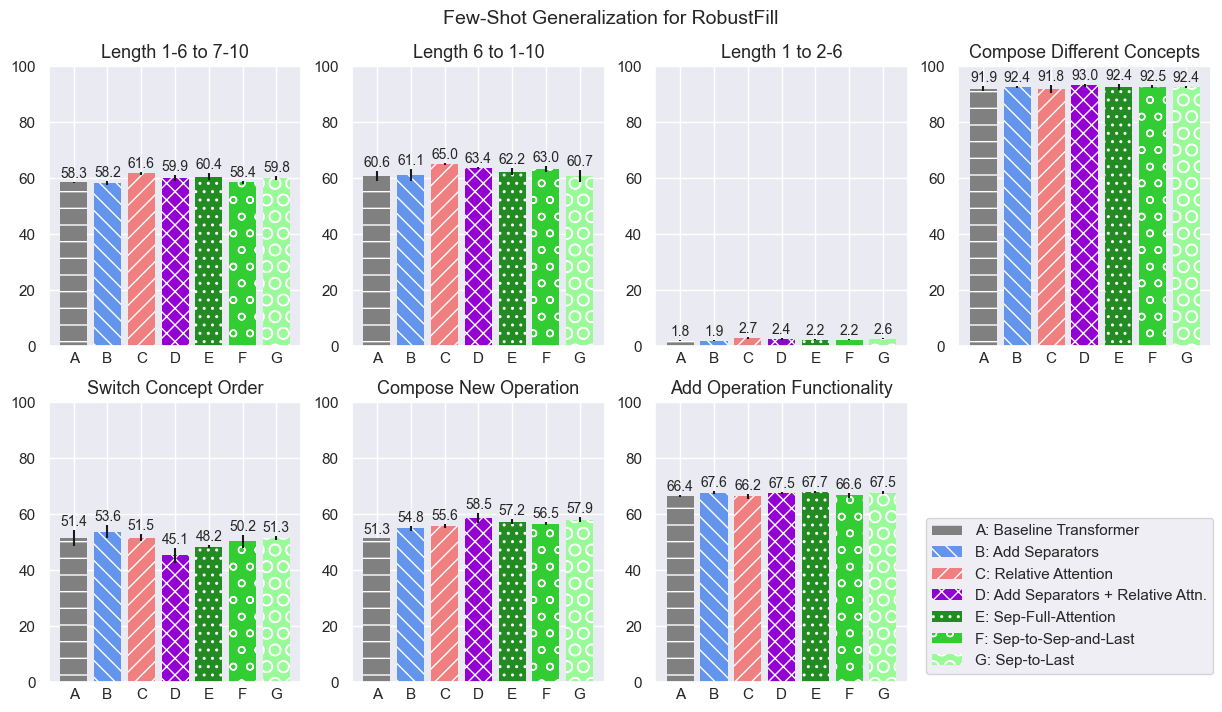}
\vspace{-2em}
\caption{Few-shot generalization for RobustFill. The same models from the zero-shot experiment were fine-tuned on a set of 20 examples from the train distribution and 20 examples from the test distribution, for 30 epochs.}
\label{fig:few-shot-robustfill}
\end{figure*}

\autoref{tab:scan} and \autoref{tab:robustfill} in \autoref{sec:experiments} provide zero-shot and few-shot generalization results for SCAN and RobustFill. \autoref{fig:zero-shot-scan}, \autoref{fig:zero-shot-robustfill}, \autoref{fig:few-shot-scan}, and \autoref{fig:few-shot-robustfill} provide bar graphs for a more visual representation of the same data.

\end{document}